\documentclass[runningheads,a4paper]{llncs}

\usepackage{amssymb}
\usepackage{graphicx}
\usepackage{caption}
\usepackage{subfig}

\usepackage{url}
\urldef{\mailsa}\path|{fzmorais, morprates, lamb}@inf.ufrgs.br|

\usepackage{cite}
\usepackage{array}
\usepackage{tikz}
\usepackage{rotating}
\usepackage{url}

\usepackage[rawfloats=true]{floatrow} 
\restylefloat{figure}     

\begin{document}

\mainmatter  

\title{Neural Networks Models for Analyzing Magic: the Gathering Cards}

\titlerunning{Neural Networks Models for Analyzing Magic: the Gathering Cards}

%
%
\author{Felipe Zilio \and Marcelo Prates \and Luis Lamb}
\authorrunning{Felipe Zilio, Marcelo Prates, Luis Lamb}

\institute{Federal University of Rio Grande do Sul, Institute of Informatics,\\
Av Bento Goncalves 9500, Porto Alegre, Brazil\\
\mailsa\\
\url{http://www.inf.ufrgs.br}}

%
%

\toctitle{Neural Networks Models for Analyzing Magic: the Gathering Cards}
\tocauthor{Felipe Zilio, Marcelo Prates, Luis Lamb}
\maketitle

\begin{abstract}
Historically, games of all kinds have often been the subject of study in scientific works of Computer Science, including the field of machine learning. By using machine learning techniques and applying them to a game with defined rules or a structured dataset, it's possible to learn and improve on the already existing techniques and methods to tackle new challenges and solve problems that are out of the ordinary. The already existing work on card games tends to focus on gameplay and card mechanics. This work aims to apply neural networks models, including Convolutional Neural Networks and Recurrent Neural Networks, in order to analyze Magic: the Gathering cards, both in terms of card text and illustrations; the card images and texts are used to train the networks in order to be able to classify them into multiple categories. The ultimate goal was to develop a methodology that could generate card text matching it to an input image, which was attained by relating the prediction values of the images and generated text across the different categories.
\end{abstract}

\section{Introduction}

Games and similar forms of entertainment, due to their constrained nature and usually tight set of rules, lend themselves especially well to be subjects of study in scientific works in the field of machine learning. 

There have been several studies applying machine learning techniques to learn and analyze several different games to great success, from traditional board games such as Go \cite{singh2017artificial} to video games including DotA 2 \cite{moreondota2017} and Starcraft II \cite{vinyals2017starcraft}. 

More recently, card games have also become a source of interest in machine learning research. In a typical card game, players are allowed to collect (or buy, or trade) several different copies of cards, with different in-game effects, which are then used to construct a ‘deck’, to play against other players. These types of games present several different angles of analysis, from the cards themselves, to deck construction, to the gameplay and its rules. 

Among the different card games, a good number of studies have focused on the bigger and most enfranchised games, such as Hearthstone \cite{zhang2017improving} and Magic: the Gathering \cite{cowling2012ensemble}. This work will focus on the latter.

Magic: the Gathering (MTG) is a fantasy-themed trading card game (TCG) created by Wizards of the Coast. Since the release of its first edition in 1993, the game has grown to be one of the biggest card games in the world in terms of number of players, with an estimated 20 million players worldwide \cite{mtgpopculture2015}, and multiple international MTG tournaments being run every year across the world. As of the time of this writing,  there are over 16 thousand unique cards with around 30 thousand images currently in the game, with more being released every few months in new editions, or sets (each set containing up to a few hundred cards, between new cards and 'reprints' of older cards, which may include new illustrations).

Although the vast number of cards in the game - each with its own different effect and mechanics - makes it an extremely complex game, the core gameplay concepts behind it are actually quite simple. The focus of this work, however, will be on analyzing the cards themselves, and not on the gameplay.


Each individual MTG card usually contains the following fields: 
\begin{itemize}
\item A name, which uniquely identifies the card. 
\item A mana cost, which is the number and colors of ‘mana’ (the game’s main resource system) required to play the card.
\item An illustration.
\item A number of types and subtypes, that define which type categories the card is a part of. 
\item The set symbol, which represents the  edition in which the card was released. 
\item Rules text defining a card’s effect.. 
\item Flavor text.
\item Power/Toughness (only on creature-type cards), values that determine the strength of a creature card.
\end{itemize}

The ‘cost’ and ‘rules text’ of a card will also determine what is called the card’s ‘color identity’. In Magic: the Gathering, a card’s color identity refers to the set of all the color symbols that appear anywhere on the card,. A card’s color identity can be any of the game’s five colors - ‘White’, ‘Blue’, ‘Black’, ‘Red’ or ‘Green’. It can also be two, three, four or five of those colors, or even none of them (‘Colorless’). Alternately, a card's color identity can be thought of as a point $x \in \{0,1\}^5$ in five-dimensional space. As such, in this work, a card’s ‘color’ will refer to its color identity as defined by the mechanics and rules of the game only.

\subsection{Color}

In Magic: the Gathering, the five-color system is one of the biggest design principles of the game. Not only does each color have its own set of effects and mechanics for its cards, but those are also related to the ideas and concepts each color represents.


Investigating the classification of card images by color is an intuitive choice since, visually, it’s easy to notice a certain correlation between the images of cards that are the same color - the tonality of an image tends to be related to the color of its card (that is, ‘Red’ cards do tend to have images with a lot of red tones in it, and so on), and the images of the cards themselves are usually created on purpose to look similar to other cards of the same color, to create a sort of ‘visual identity’ for each color through the card images. The similarities between cards of a certain color also extend to the cards' rules text - since there is a clear mechanical and gameplay separation for each different card type, the words used and even the way the card's effect is structured will usually be different for each card type. This distinction also extends to the cards' color identity, albeit in a more subtle manner; just as each color has a different visual theme for its images, the gampeplay mechanics and effects of each color also tends to differ - that is, there is a mechanical, or gameplay, identity for each color that tends to appear in the cards' text. As such, since this definition will remain relevant to the rest of this work, this section will provide a brief description of each of the game’s colors, both in terms of card art and game mechanics, taken directly from a series of articles on the subject by one of the game’s developers \cite{colorpie2004}.


\begin{table}
  \centering
  \begin{tabular}{|c|m{3.5cm}|m{3.5cm}|}
  \hline
  Color   & Attributes                                & Common Elements in Illustrations \\ \hline \hline
  White   & Order, Structure, Law                     & Soldiers, Angels, Clerics, Light, The Sun \\ \hline
  Blue    & Intellect, Manipulation, Cold, Water, Air & Books, Wizards, Clouds, The Sky, Bodies of Water, Birds, Sea Creatures \\ \hline
  Black   & Ambition, Amorality, Sacrifice, Death     & Zombies, Demons, Vampires, Darkness \\ \hline
  Green   & Nature, Life, Growth                      & Animals, Plants, Elves \\ \hline
  Red     & Impulse, Chaos, Earth, Fire, Lightning    & Fire, Aggression, Lightning, Dragons, Goblins \\ \hline
  \end{tabular}
  \caption{Conceptual attributes and common elements associated with each card color.}
\end{table}

\subsection{Card Type}

Each card in Magic: the Gathering belongs to one or more card types. The main card types - those being \textbf{Land}, \textbf{Creature}, \textbf{Instant}, \textbf{Sorcery}, \textbf{Artifact}, \textbf{Enchantment} and \textbf{Planeswalker} -- are already very descriptive: Land cards are places or landscapes, Creature cards are beings like beasts or humanoids, Instant and Sorcery cards represent magical spells, Artifacts are inanimate objects, Enchantments are long lasting changes to the battlefield or other enhancements, Planeswalker cards represent specific named wizards that are part of the game's story. More secondary card types exist, but any card in the game belongs to at least one of the previously mentioned card types.

\section{Our Goals}
Magic: the Gathering cards are comprised of two elements: the rules text (including the textbox as well as its name, cost, and other relevant information printed on the card), and its illustration. Previous projects on the subject of this card game have mostly focused on a card’s text and its in-game mechanics, such as the card-text generating neural network RoboRosewater, which posts the results of its generated cards on Twitter \cite{mtgroborosewater}, and Google DeepMind’s project for generating programming code to implement cards through the use of latent prediction networks \cite{ling2016latent}.

In this work, the focus will be on the cards’ illustrations, and how they relate to elements of the game such as the different categories of cards, and the card text itself. A process will be described to classify card images according to different types of card characteristics, which will then be used to relate input images to randomly generated card texts according to those parameters, learned from training with a dataset containing all of the game’s card images.

\section{Methods}

Card data was pulled directly from Wizards of the Coast’s database \cite{mtggatherer}, containing all the current cards in the game. All card images were downloaded in JPG format, and all card text information in CSV format. All the images were converted to binary format, and encoded in batch files along with the relevant label for that dataset. Two datasets were created, one for card type classification, one for color identity classification, of encoded images and labels separated in batches. These datasets were then fed to a convolutional neural network, configured slightly differently for each classification, the specifics of which will be detailed in later sections of this work. Another neural network was used to generate random cards from the dataset of existing card text already in the game. These cards were used to form a database of 20 thousand  generated cards, to which input images could then be related according to their characteristics.

All card images experimented with in this study are suppressed from the manuscript due to copyright restraints. Nevertheless, they can be obtained by typing the card names in the search engine \url{http://gatherer.wizards.com}.

\subsection{Environment}
All experiments were conducted in a machine with the specifications: i5-4690k quad-core CPU @ 3.5 GHz, NVIDIA GTX 970 graphics card with 4GB VRAM, 16 GB of RAM, running the Ubuntu 10.4 OS.

\section{Classifying MTG Card Illustrations by Color}
\label{sec:classifying_by_color}

Out of the 32.000 images in the dataset, about 26.000 images were used during the training. The rest was randomly selected to form the test batch of images, for evaluation purposes. This ensures a good mix of different image styles from all different eras and editions of Magic: the Gathering history, for both training and testing. A set amount of images was also distorted at random through cropping and displacing, a method which has been shown to increase the performance of the network in the past \cite{simard2003best} and was effective here as well, offering a slight gain in performance when compared to a training session that was run with just the regular images as inputs.

The convolutional neural network architecture that was used for this task was adapted from a CNN model \cite{cudaconvnet2011} for image classification. The network was created through the use of consecutive convolution, pooling, normalization, and softmax layers, connected to form the network structure.

For the network parameters, a small initial learning rate of 0.01 was used - higher initial values usually resulted in the network diverging.

Our neural network is composed of 8 convolution, pooling, normalization, fully connected and softmax layers arranged in the following way: \textbf{CONV} $\rightarrow$ \textbf{POOL} $\rightarrow$ \textbf{NORM} $\rightarrow$ \textbf{CONV} $\rightarrow$ \textbf{NORM} $\rightarrow$ \textbf{POOL} $\rightarrow$ \textbf{FULLY} $\rightarrow$ \textbf{SOFTMAX}.

\ifx
\begin{table}
  \centering
  \begin{tabular}{|c|c|c|c|}
  \hline
  Type                        & Input Dimension           \\ \hline \hline
  Convolutional               & $W \times H \times 3$     \\ \hline
  Pooling                     & $(W-5) \times (H-5)$      \\ \hline
  Normalization               & $(W-5)/3 \times (H-5)/3$  \\ \hline
  Convolutional               &                           \\ \hline
  Normalization               &                           \\ \hline
  Pooling                     &                           \\ \hline
  Fully Connected             &                           \\ \hline
  Softmax                     &                           \\ \hline
  \end{tabular}
  \caption{conv 5x5 stride 1 ReLu}
\end{table}
\fi

Table ~\ref{tbl:colordist-tbl} shows the distribution of color identity across the $32.017$ cards in the dataset. There are 3824 multicolored cards, or 11.94\%. It would be extremely hard to classify card images across all color combinations using single labels: not only are there $2^5 = 32$ possible combinations, from ‘Colorless’ to ‘All colors’, but there is a big imbalance across each category in terms of numbers of cards. In fact, while there are $5.068$ Green cards in total, there are only two cards currently in the game in the White/Blue/Red/Green color identity, for example.

\begin{table}[H]
\caption{Color distribution in the dataset - number of images}
\centering
    \begin{tabular}{| l | l | l | l |}
    \hline
    Color & Images & Color & Images \\ \hline \hline
    G & 5068 & UBR & 99\\ \hline
    B & 5021 & WRG & 86\\ \hline
    R & 4973 & WUG & 79\\ \hline
    W & 4878 & BRG & 79\\ \hline
    U & 4828 & WUB & 75\\ \hline
    Cl & 3425 & WUBRG & 66\\ \hline
    WG & 378 & WBR & 49\\ \hline
    BR & 376 & WBG & 41\\ \hline
    RG & 372 & UBG & 38\\ \hline
    UB & 354 & WUR & 37\\ \hline
    WU & 346 & URG & 37\\ \hline
    BG & 288 & UBRG & 2\\ \hline
    WB & 266 & WUBR & 2\\ \hline
    UR & 260 & WUBG & 2\\ \hline
    WR & 249 & WBRG & 2\\ \hline
    UG & 238 & WURG & 2\\ \hline
    \end{tabular}
\label{tbl:colordist-tbl}
\end{table}

While imbalanced categories are common problems with image datasets, and can be usually addressed with simple solutions such as re-weighting the training cost function or resampling the training images in order to balance the dataset \cite{oquab2014learning}, in this case, the imbalance present in the dataset was just too big for these solutions to be considered feasible. Sticking to 6 labels, then - one for each color, plus colorless - was the system that was used for initially classifying the dataset. 

We can visually verify that multicolored cards do usually contain traces of all colors in their color identity in their illustration picture -- a card that was both Red and Blue, for example, would usually present characteristic exhibited by both Red and Blue cards in its picture. As such, we have decided to include a copy of each multicolored card in the dataset of each of its respective colors.


The final accuracy of the network, when trained with the merged dataset, was 0.595, or 59.5\%. This value, while not necessarily low, might possibly be explained at least in part by the huge variance in image styles that are present in the dataset, as well as the fact that different eras of Magic: the Gathering cards present different ideas and sometimes completely different styles of cards when it comes to color, since the color system is always in flux as the game develops \cite{colorpie2017}.

Consequently, it needs to be taken into account, then, that at least some of the prediction ‘misses’ will be the result of images from early in the game’s history, some of which are unfitting for their color identity as per today’s Magic: the Gathering card design standards. Table \ref{tbl:img-color} shows an example of prediction values obtained for a multicolored card.


\begin{table}[H]
\caption{Prediction values for card color}
\centering
    \begin{tabular}{| l | l |}
    \hline
    Type & Prediction Value \\ \hline \hline
    White & 27,49\% \\ \hline
    Green & 27,15\% \\ \hline
    Black & 27,14\% \\ \hline
    Blue & 9,73\% \\ \hline
    Red & 8,49\% \\ \hline
    Colorless & 0,00\% \\ \hline
    \end{tabular}
\label{tbl:img-color}
\end{table}

\section{Classifying MTG Card Illustrations by Type}
\label{sec:classifying_by_type}

Similarly to color classification, classifying cards by type is also an intuitive notion. Since the cards are already pre-labeled with their type, and each card type does share a few similarities and characteristics between its images, it stands to reason that using that criteria for classification would be an effective choice. 

Card type as a method of classification also presents a much clearer correlation to images outside of the context of the game, when compared to the color classification, since the characteristics of the images of each type are much more well defined (‘creatures’ are living things, ‘artifacts’ are inanimate objects, ‘lands’ are landscapes or places, etc). Table \ref{tbl:typedist-tbl} shows the distribution of cards among all types.

\begin{table}[H]
    \caption{Type distribution in the dataset - number of images}
    \centering
    \begin{tabular}{|c|c|}
    \hline
    Type                    & \# Cards  \\ \hline \hline
    Creature                & 14081 \\ \hline
    Instant                 & 3962  \\ \hline
    Sorcery                 & 3638  \\ \hline
    Enchantment             & 3519  \\ \hline
    Land                    & 3297  \\ \hline
    Artifact                & 2259  \\ \hline
    Creature/Artifact       & 920   \\ \hline
    Planeswalker            & 179   \\ \hline
    Creature/Enchantment    & 98    \\ \hline
    Land/Artifact           & 14    \\ \hline
    Instant/Sorcery         & 13    \\ \hline
    Artifact/Enchantment    & 8     \\ \hline
    Creature/Land           & 2     \\ \hline
    \end{tabular}
    \label{tbl:typedist-tbl}
\end{table}

This classification system suffers from the exact same problem as the color system; that is, there are cards that belong to more than one type category(for example, Enchantment Creature or Artifact Creature). The solution described before for the case of multicolor cards was also proposed and utilized here, since, in this case as well, images of cards that have multiple types are usually representative of both types. We have also decided to merge 'Instant' and 'Sorcery' types into a single category, given the poor differentiation between both types in the philosophy of the game.

The implemented CNN was trained with this dataset, and it was able to obtain an accuracy of 0.685, or 68.5\%, in its predictions - a marked increase in performance when compared to the color classification.


Table ~\ref{tbl:img-type} shows an example of prediction values obtained from evaluating the card art in the 2011 core set Ornitopher card with the trained type classification CNN.

\begin{table}[H]
\caption{Prediction values for card type}
\centering
    \begin{tabular}{| l | l |}
    \hline
    Type            & Prediction Value  \\ \hline
    Creature        & 44,06\%           \\ \hline
    Artifact        & 43,82\%           \\ \hline
    Enchantment     & 7,95\%            \\ \hline
    Instant/Sorcery & 4,17\%            \\ \hline
    Land            & 0,00\%            \\ \hline
    \end{tabular}
\label{tbl:img-type}
\end{table}

\section{Text Classification}
\label{sec:text_classification}

Another common problem studied in the field of machine learning is that of text classification. Previous works on the subject of text analysis have studied methods used to classify text according to anything from sentiment analysis \cite{maas2011learning}, to even more practical applications such as spam filtering \cite{sahami1998bayesian}. Convolutional Neural Networks have been used in the past to solve natural language processing tasks of various sorts \cite{collobert2011natural}.

The method used in this work embeds each piece of text into a sparse vector representations \cite{kim2014convolutional} which can then be used as inputs to be passed to the CNN on the training or evaluation steps. The data is organized much like on the experiments described in the previous section, with the same categories for color and card type. The complete dataset was split into a training dataset and an evaluation dataset, with a ratio of 5/6 for training to 1/6 for evaluation. The overall method utilized for the task of text classification in this work differs slightly from the reference model in \cite{kim2014convolutional}, but the network structure is otherwise very similar. After the preliminary embedding layer, the data is passed through the usual convolution, pooling and fully connected layers, providing the output through the result of a softmax function. Although the details of this architecture differ heavily from the network used previously for the task of image classification, the general idea of each layer remains mostly the same and as such will not be repeated in detail in this section.

Given the relative simplicity of card text when compared to card images, the overall better results obtained from this classification experiment, when compared to the image classification, are not surprising.  The CNN used to classify the color dataset was able to predict categories with 0.78 accuracy, and the second dataset, used for type classification, yielded a prediction accuracy of 0.91. Tables ~\ref{tbl:text-type} and ~\ref{tbl:text-color}  shows an example of text classification done through the network, for card type and color, on the text of the ``Journey into Nyx'' expansion ``Kruphix, God of Horizons'' card.


\begin{table}
\caption{Prediction values for card type}
\centering
    \begin{tabular}{| l | l |}
    \hline
    Type & Prediction Value \\ \hline
    Enchantment & 37,65\% \\ \hline
    Creature & 36,44\% \\ \hline
    Artifact & 18,44\% \\ \hline
    Instant/Sorcery & 7,48\%\\ \hline
    Land & 0,00\% \\ \hline
    \end{tabular}
\label{tbl:text-type}
\end{table}

\begin{table}
\caption{Prediction values for card color}
\centering
    \begin{tabular}{| l | l |}
    \hline
    Type & Prediction Value \\ \hline
    Green & 27,34\% \\ \hline
    Blue & 26,37\% \\ \hline
    White & 17,79\% \\ \hline
    Black & 16,34\% \\ \hline
    Red & 12,16\% \\ \hline
    Colorless & 0,00\% \\ \hline
    \end{tabular}
\label{tbl:text-color}
\end{table}{}

\section{Matching Randomly Generated Card Text to Input Images}

As a part of this research endeavor, we used the prediction values for card color and type obtained from the CNNs for image and text classification described in Sections \ref{sec:classifying_by_color} and \ref{sec:classifying_by_type} in order to develop a method that can relate an input image to a randomly generated card text that better fits it -- effectively automatically “creating” a Magic: the Gathering card for any input image. A few of these examples are described above.

All randomly generated card text used in this work was entirely generated through the method described in \cite{mtgroborosewater}, which contains a set of methods to encode the dataset of MTG cards in JSON format to be used as input. Then, it utilizes a Recurrent Neural Network model to generate a set number of examples from the training data. The card generating RNN was used to generate a database of $30.000$ card text data. This card text was then processed through the text classifying CNN developed previously on Section \ref{sec:text_classification}, to classify the generated text according to both type and color.

An input image will first be run through two CNNs -- the first one, described in section \ref{sec:classifying_by_color}, will classify the image according to its potential color identity, and return the prediction values obtained for each possible color label. The same process will be repeated for the second CNN, on section \ref{sec:classifying_by_type}, to obtain prediction values for each card type label.

After a normalization process these values can be directly compared to the prediction values previously obtained and recorded for each generated card. This comparison is a simple sum of the differences between the value of each label for the two categories, in order to compute the total color and type ‘distance’ $ C_{d} $ and $ T_{d} $: $C_{d}  = \sum_{{c_i} \in C} ~ |I_{c_i} - T_{c_i}|$, where $c_i$ represents the possible color labels in the space of color labels $C$ , $I_{c_i}$ is the prediction value of the color label $ci$, obtained by the CNN for the image $I$, and $T_{c_i}$ is the prediction value of the color label $c_i$ on the generated card text $T$. Likewise, for obtaining the type distance $T_{d}$, the same formula can be applied, by utilizing the type labels instead of the color ones.   

Finding the best possible match between the predicted color and type values for the input image and the generated card texts becomes a simple matter of finding the card text that minimizes both distance values $C_{d}$ and $T_{d}$ - this selected card text will then will be the sole output of the method. Some examples of our results are described above.

\subsection{Card Generation Examples}

\begin{figure}%
    \centering
    \subfloat{{\includegraphics[width=.4\linewidth]{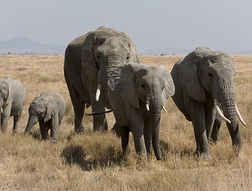} } \label{fig:elephants}}%
    \qquad
    \subfloat{{\includegraphics[width=.4\linewidth]{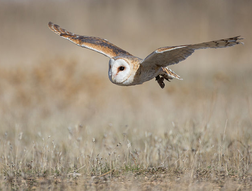} } \label{fig:owl}}%
    \caption{Generated card examples 1 and 2}%
    \label{fig:elephant_owl}
\end{figure}

Image ~\ref{fig:elephants} shows a herd of elephants. The prediction values for type and color, on table ~\ref{tbl:pred-elephants},  show a strong preference for the Creature type, and Green color, closely followed by Red. This decision is consistent with the philosophy of the Green color, which is home to most animals and beasts. The generated card text (\emph{"Cumulative upkeep {1}. At the beginning of your upkeep, you may pay {3G}. if you do, put a +1/+1 counter on it."}) both fits the description of a Green Creature as well as being a coherent card effect. 

\hfill \\

\begin{table}[H]
\caption{Color and type prediction values for examples 1}
\centering
    \begin{tabular}{| l | l | l | l |}
    \hline
    Color & Color Values  & Type & Type Values\\ \hline \hline
    Green & 27,22\% & Creature & 39,73\% \\ \hline
    Red & 26,02\% & Enchantment & 27,89\% \\ \hline
    White & 17,19\% & Artifact & 22,97\% \\ \hline
    Black & 17,17\% & Instant/Sorcery & 9,41\% \\ \hline
    Colorless & 12,39\% & Land & 0,00\%\\ \hline
    Blue & 0,00\% & - & - \\ \hline
    \end{tabular}
\label{tbl:pred-elephants}
\end{table}

\begin{table}[H]
\caption{Color and type prediction values for example 2}
\centering
    \begin{tabular}{| l | l | l | l |}
    \hline
    Color & Color Values  & Type & Type Values\\ \hline \hline
    Blue & 29,49\% & Creature & 55,11\% \\ \hline
    White & 29,03\% & Artifact & 26,66\% \\ \hline
    Black & 15,80\% & Instant/Sorcery & 12,87\% \\ \hline
    Red & 14,65\% & Enchantment & -5,36\% \\ \hline
    Green & 11,04\% & Land & 0,00\%  \\ \hline
    Colorless & 0,00\% & - & -\\ \hline    
    \end{tabular}
\label{tbl:pred-owl}
\end{table}

Image ~\ref{fig:owl} shows an owl mid-flight. The values predicted by the CNN for this image are reported on table ~\ref{tbl:pred-owl}. Once again the NN easily recognizes it as Creature type card. Color-wise, strong values of Blue and White are predicted. The fact that in MTG birds of all kinds are very common on both colors probably explains the Blue classification even though the image lacks Blue hues.

The card text generated (\emph{"Flying. When \{this card\} enters the battlefield, detain target creature an opponent controls."}), again, is correct by the rules of the game, and features the already existing 'Detain' ability, which shows up exclusively in Blue/White cards. The card having the ability 'Flying' is interesting, since it's not possible to identify that it is an aerial creature only by the 'Blue/White Creature' description, which is all the network has.

\section{Discussion and Future Work}

The main objective of this work was to assess whether state-of-the-art convolutional neural network architectures could tackle a classification problem that remained up to now untouched -- that is, the dataset of a major trading card game. The image classification problem presented an interesting challenge in regards to the usual image classification problems: instead of labelling images based on elements that were or not present, a more subjective classification system was employed, based on game rules and mechanics not necessarily related to the images themselves.

The relative success of the classification phase presented an opportunity to test a hypothesis -- that it would possible to relate the values of predictions obtained on the images for the different categories, with the values obtained from a classification operation on the card text, in a way that would enable the approximation of a given image to a text that most matches it. Thus, the second CNN model was utilized, to classify existing card text in order to train a network to be able to classify newly generated card text. This classification yielded even better results. A model for generating random card text, based on a RNN model, was then utilized, and each of its cards were classified with the already trained text classification CNN. This allowed for a direct comparison to the prediction values obtained from the input image to be made.

The method developed was tested with real-world images, from outside the game. The generality of the categories used, color identity and card type, allowed images that had nothing to do with the game to be analyzed and classified according to the same criteria used for card images in the game -- this stems from the intuitive notion that, for example, if the Creature card type is comprised of animals and creatures, then an image of a real animal, if recognized, should fit into the same classification. The testing process, while not extensive, already yielded a few significant insights and conclusions.

Overall, the results of this work present an exciting possibility for future research -- since there are so many independent components to this work, from the classifying CNNs to the generating RNN, there are multiple areas that could be improved upon to make the task of generating cards to match input images yield better results. The obvious two candidates are the image classification and text generation phases, which can be improved on multiple fronts. Also, there still lacks a formal, extensive review of the effectiveness of the methodology utilized -- one possible idea is to conduct a quantitative research, asking Magic: the Gathering players to evaluate the quality of the generated cards on a certain scale.

\bibliographystyle{splncs03}
\bibliography{main}

\end{document}